# Symbolic and Language Agnostic Large Language Models

Walid S. Saba[†]
Institute for Experiential AI, Northeastern University
Portland, ME, USA
w.saba@northeastern.edu

## ABSTRACT

We argue that the relative success of large language models (LLMs) is not a reflection on the symbolic vs. subsymbolic debate but a reflection on employing an appropriate strategy of bottom-up reverse engineering of language at scale. However, due to the subsymbolic nature of these models whatever knowledge these systems acquire about language will always be buried in millions of microfeatures (weights) none of which is meaningful on its own. Moreover, and due to their stochastic nature, these models will often fail in capturing various inferential aspects that are prevalent in natural language. What we suggest here is employing the successful bottom-up strategy in a symbolic setting, producing symbolic, language agnostic and ontologically grounded large language models.

## CCS CONCEPTS

• Insert CCS text here

## 1 Introduction

To arrive at a scientific explanation there are generally two approaches we can adopt, a top-down approach or a bottom-up approach [1]. However, for a top-down approach to work, there must be a set of established general principles that one can start with, which is clearly not the case when it comes to language and how our minds externalize our thoughts in language. In retrospect, therefore, it is not surprising that decades of top-down work in natural language processing (NLP) failed to produce satisfactory results since most of this work was inspired by theories that made assumptions that were controversial, to say the least (e.g., [2] where an innate universal grammar was assumed, or [3] where it was postulated that we 'metaphorically' build our linguistic competence based on a set of idealized cognitive models, or [4] where it was postulated that natural language could be formally described using the tools of formal logic). In a similar vein, it is perhaps for the same reason that decades of work in ontology and knowledge representation (e.g., [6] and [7]) also faltered since most of this work amounted to pushing, in a top-down manner, metaphysical theories of how the world is supposedly structured and represented in our minds, and again without any agreed upon general principles to start with. On the other hand, unprecedented progress has been made in only a few years of NLP work that employed a bottom-up strategy, as exemplified by recent advances in large language models (LLMs) that are essentially a massive experiment of a bottom-up reverse engineering of language at scale (e.g., ChatGPT and GPT-4)[1].

Despite their relative success, however, LLMs do not tell us anything about how language works since these models are not really *models of language* but are statistical models of regularities found in linguistic communication. In fact, and due to their subsymbolic nature, whatever 'knowledge' these models acquire about language will always be buried in millions of weights (microfeatures) none of which is meaningful on its own, rendering these models to be utterly unexplainable[2]. Besides their lack of explainability, LLMs are oblivious to truth, since for LLMs all ingested text is treated equally, whether factual or non-factual. Another related issue is that LLMs are not grounded in any ontological reality, and they will therefore often fail in making the right inferences that require commonsense reasoning in complex intensional, modal, and temporal contexts. Consider for example the following exchange with GPT-4:

PROMPT    What is 'informed' modifying here?
              Mary gave a very informed talk at AAAI.

LLM    In the sentence "Mary gave a very informed talk at AAAI," the word 'informed' is modifying the noun 'talk.'

Note that the LLM wrongly applies the modifier to 'talk' based solely on the surface (syntactic) structure, although a 'talk' (an abstract object/an event) cannot be 'informed' but it is the person who gave the talk that is presumed to be informed here. These inferential mistakes, while seemingly trivial, make LLMs unusable in many critical applications (e.g. querying enterprise databases). As another example, consider the following:

PROMPT    Sara who is from NY City, met John who is the mayor of a small city in California, when they were both on vacation in Paris. They had a beautiful time and became close friends. Now Sara keeps telling her friends that she

---

[1] GPT stands for 'Generative Pre-trained Transformer', an architecture that OpenAI built on top of the transformer architecture introduced in [5].

[2] Explainability is 'inference in reverse' and thus a computation is explainable if (i) the computation is invertible; or (ii) a semantic map of the computation is saved in some symbolic structure (e.g., an abstract syntax tree), and neither of these apply to subsymbolic deep neural networks [8].

> hopes to see John again and relive those beautiful two weeks. Based on this, how much time did Sara and John spend in Paris?

LLM  The information provided only mentions that Sara and John had a beautiful time and became very close friends during their vacation in Paris. The duration of their stay is not mentioned, so we cannot determine how much time they spent in Paris based on the given information.

The LLM clearly fails to resolve 'those beautiful two weeks' with the time Sara and John spent in Paris. Again, these failures preclude using LLMs in mission critical applications where factual information and valid inferences are of paramount importance.
So where do we stand now? On one hand, LLMs have clearly proven that one can get a handle on syntax and quite a bit of semantics in a bottom-up reverse engineering of language at scale; yet on the other hand what we have are unexplainable models that do not shed any light on how language actually works and how we externalize our thoughts in language. Moreover, it would seem that due to their stochastic nature, LLMs will always fail in making the correct inferences that are often implicit in linguistic communication. Since we believe the relative success of LLMs is not a reflection on the symbolic vs. subsymbolic debate but is a reflection on a successful bottom-up reverse engineering strategy, we think that combining the advantages of symbolic and ontologically grounded representations with a bottom-up reverse engineering strategy is a worthwhile effort. In fact, the idea that word meaning can be extracted from how words are actually used in language is not exclusive to linguistic work in the empirical tradition, but in fact it can be traced back to Frege, although there are more recent philosophical and computational proposals on what this project amounts to. Below we discuss these proposals and suggest how this project can be accomplished.

## 2 Concerning 'the Company a Word Keeps'

The genesis of modern LLMs is the *distributional semantics hypothesis* which states that the more semantically similar words are, the more they tend to occur in similar contexts – or, similarity in meaning is similarity in linguistic distribution [9]. This is usually summarized by a saying that is attributed to the British linguist John R. Firth that "you shall know a word by the company it keeps". When processing a large corpus, this idea can be used by analyzing co-occurrences and contexts of use to approximate word meanings by word embeddings (vectors), that are essentially points in multidimensional space. Thus, at the root of LLMs is a bottom-up reverse engineering of language strategy where, unlike top-down approaches, "reverse engineers the process and induces semantic representations from contexts of use" (see [10] for more details). But nothing precludes this ingenious idea from being carried out in a symbolic setting. In other words, the 'company a word keeps' can be measured in several ways, other than the correlational and statistical measures that underlie modern word embeddings. For example, in discussing possible models (or theories) of the world that can be employed in computational linguistics Jerry Hobbs [11] once suggested that there are two alternatives: on one extreme we could attempt building a "correct" theory that would entail a full description of the world, something that would involve quantum physics and all the sciences; on the other hand, we could have a promiscuous model of the world that is isomorphic to the way we talk it about in natural language. Clearly, what Hobbs is suggesting here is a reverse engineering of language to discover how we actually use language to talk about the world we live in. In essence, this is not much different from Frege's Context Principal that suggests to "never ask for the meaning of words in isolation" [12] but that a word gets its meanings from analyzing all the contexts in which the word can appear [13].

Again, what this suggests is that the meaning of words is embedded (to use a modern terminology) in all the ways we use these words in how we talk about the world. While Hobbs' and Frege's observations might be a bit vague, the proposal put forth by Fred Sommers [14] was very specific. Again, Sommers suggests that "to know the meaning of a word is to know how to formulate some sentences containing the word" and this would lead, like in Frege's case, to the conclusion that a complete knowledge of some word $w$ would be all the ways $w$ can be used. For Sommers, the process of understanding the meaning of some word $w$ starts by analyzing all the properties $P$ that can sensibly be said of $w$. Thus, for example, [*delicious Thursday*] is not sensible while [*delicious apple*] is, regardless of the truth or falsity of the predication. Moreover, and since [*delicious cake*] is also sensible, there must be a common type (perhaps food?) that subsumes both *apple* and *cake*. This idea is similar to the idea of type checking in strongly typed polymorphic programming languages. For example, the types in an expression such as '$x + 3$' will only unify (or the expression will only 'make sense') if/when $x$ is an object of type number (as opposed to a tuple, for example). As it was suggested in [15], this type of analysis can be used to 'discover' the ontology that seems to be implicit in all natural languages. What we suggest here is to generate all sets $Cp = \{c \mid \mathbf{app}(p, c)\}$ for all concepts (nouns) $c \in C$ and all properties (relations) $p \in P$ such that the property $p$ is applicable to (can sensibly be said of) $c$. The following are examples of successful applications of **app**($p$, $c$):

$R_1$ : **app**(OLD, entity)
$R_2$ : **app**(HEAVY, physical)
$R_3$ : **app**(HUNGRY, living)
$R_4$ : **app**(ARTICULATE, human)
$R_5$ : **app**(MAKE(human, artifact))
$R_6$ : **app**(MANUFACTURE(human, instrument))
$R_7$ : **app**(RIDE(human, vehicle))
$R_8$ : **app**(DRIVE(human, car))

What the above say, respectively, is the following:

$R_1 \rightarrow$ in ordinary language we can say OLD of any entity
$R_2 \rightarrow$ we say HEAVY of objects that are of type physical
$R_3 \rightarrow$ HUNGRY is said of objects that are of type living
$R_4 \rightarrow$ ARTICULATE is said of objects that are of type human
$R_5 \rightarrow$ MAKE holds between a human and an artifact
$R_6 \rightarrow$ MANUFACTURE relates a human and an instrument
$R_7 \rightarrow$ RIDE holds between a human and a vehicle
$R_8 \rightarrow$ DRIVE holds between a human and a car

Note that the above 'findings' would eventually result in a well-defined hierarchy. For example, since a bottom-up reverse engineering of language will ultimately produce **app**(HEAVY, car) and **app**(OLD, car) – that is, since by analyzing our linguistic communication, we would also discover that it is sensible to say 'heavy car' and 'old car' it would seem that car must be a subtype of physical which in turn must be a subtype of entity. Similarly, since it makes sense to say MAKE of everything we MANUFACTURE, an instrument must be a subtype of artifact. The fragment of the hierarchy that is implicit in $R_1$ through $R_8$ is shown in figure 1.
**Figure 1.** The hierarchy implicit in the 'discoveries' $R_1$ through $R_8$ above.

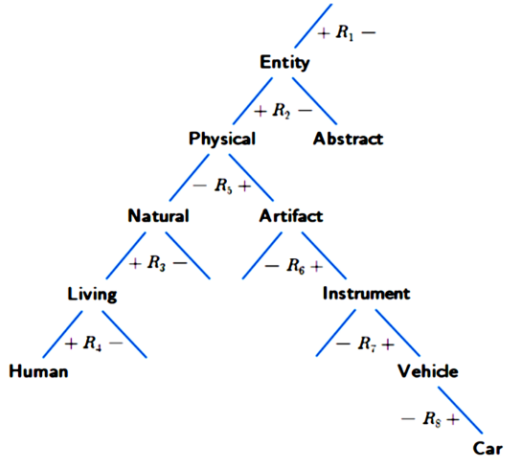

Note, also, since **app**(ARTICULATE, human) says that 'articulate' is a property that can be said of objects of type human, we can rewrite this fact as **hasProp**(articulation, human), where ARTICULATE is reified (nominalized) as the trope articulation which is an abstract object of type property (see [16] for more details on such abstract objects). Using the primitive and linguistically agnostic relation **hasProp** what we now have is a relation between two entities, the property of articulation and a human, which effectively states that articulation is a property that is usually attributed to (or said of) objects that are of type human. The same can be done with $R_3$, **app**(HUNGRY, living), resulting in **inState**(hunger, living) that says any living entity can be in a state of hunger. The result of this discovery process (that produces linguistic knowledge such as $R_1$ through $R_8$) coupled with the nominalization process and using primitive relations between entities will be no less than discovering (as opposed to inventing) the ontology that seems to be implicit in ordinary language. The details of how the *language agnostic primitive relations* can be discovered requires more space than we have here, but table 1 shows how some of these relations can be extracted from how different types of objects are described in ordinary language.

Here's a summary of the overall process: (*i*) by analyzing a large corpus we can discover all pairs of *c* and *p* for which **app**(*p*, *c*) holds (e.g., **app**(ARTICULATE, human), **app**(HUNGRY, living)); (*ii*) via a nominalization process convert **app**(*p*, *c*) to two entities related by some primitive relation (e.g., **hasProp**(human, articulation), **inState** (human, illness)); (*iii*) construct the ontology implicit in all the discovered relations.

## 3  Dimensions of Word Meaning

What we have suggested thus far is a bottom-up reverse engineering of language using the predicate **app**(*p*, *c*) that effectively generates sets for all nouns *c* that the property *p* is applicable of. This in turn can be converted into a triple **R**(entity, entity) corresponding to([entity] → (**R**) → [entity]) after all the concepts have been reified, where **R** is a primitive and language agnostic relation. Since every entity can now be defined by primitive relations, these primitive relations would now represent what we call *the dimensions of word meanings*. In figure 2 we show these dimensions for (one of) the meanings of the word book.

**Table 1.** Discovering the language agnostic primitive relations.

| LINGUISTIC CONTEXT | IMPLICIT PRIMITIVE RELATION |
| --- | --- |
| Frido *is* a dog | Frido *instanceOf* dog |
| Billy the Kid *is* William H. Boney | Billy the Kid *eq* William H. Boney |
| JFK *is* John Fitzgerald Kennedy | JFK *eq* John Fitzgerald Kennedy |
| Mary *is* wise | Mary *hasProp* wisdom |
| Julie *is* articulate | Julie *hasProp* articulation |
| Jim *is* sad | Jim *inState* sadness |
| Carlos *is* ill | Carlos *inState* illness |
| Sara *is* running | Sara *agentOf* running |
| Olga *is* dancing | John *agentOf* dancing |
| Sara *is* greeted | Sara *objectOf* greeting |
| Sara *is* acknowledged | Sara *objectOf* acknowledgment |
| John *is* 5'10'' tall | John's *height hasValue* 5'10'' |
| Dan *is* 69 years old | Dan's *age hasValue* 69 yrs |
| Sheba *is* running | Sheba *participantIn* running (event) |
| Olga *is* dancing | Olga *agentOf* dancing (activity) |

As shown in figure 2, one meaning of the word *book* (namely "a written work or composition that has been published") is an entity (*i*) that can be the agent of a changing event (as in '*Das Kapital changed many opinions on capitalism*'); (*ii*) that can have the popularity property (as in '*The Prince is a popular book*'); (*iii*) that can be the object of an inspiring event (as in '*Hamlet inspired many movies*').

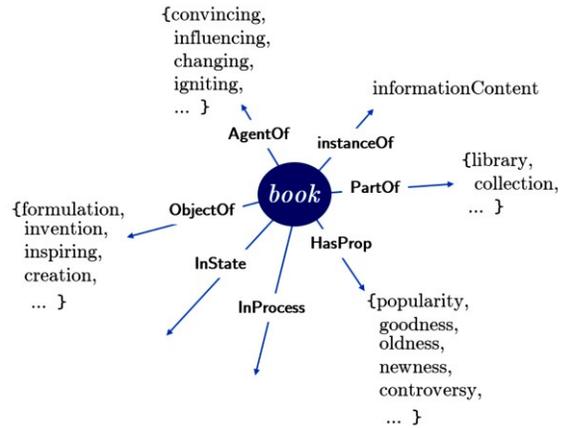

**Figure 2.** The primitive and linguistically agnostic relations as the dimensions of word meaning (in this case of one meaning of 'book').

We can use GPT-4 to generate some of these vectors along the various dimensions. The data in figure 3 is obtained by asking GPT-4 to provide 25 "plausible" (or "sensible") replacements for the [**MASK**] in the given sentences along three dimensions, namely **AgnetOf**, **ObjectOf** and **HasProp**, respectively.

Note that using this strategy we can also 'discover' the underlying ontology that seems to be implicit in our ordinary language. In figure 3 we apply masking to generate the most plausible actions that a computer, a car, and a couch can be the object of. Note that while the three types of objects can be the objects of ASSEMBLE (we 'assemble' computers, cars, and couches), we can sensibly say a computer or a car is RUNNING (or that a computer or a car is OFF) but the same is not true of a couch. This tells us that while a computer and a car must have some common supertype (say a machine), these two types seem to eventually belong to a different branch from couch although the three objects must have a common supertype at some level of abstraction since they can all be 'assembled' (see figure 4).

figure 3 we apply masking to generate the most plausible actions that a computer, a car, and a couch can be the object of. Note that while the three types of objects can be the objects of ASSEMBLE (we 'assemble' computers, cars, and couches), we can sensibly say a computer or a car is RUNNING (or that a computer or a car is OFF) but the same is not true of a couch. This tells us that while a computer and a car must have some common supertype (say a machine), these two types seem to eventually belong to a different branch from couch although the three objects must have a common supertype at some level of abstraction since they can all be 'assembled' (see figure 4).

## 4 Concluding Remarks

Large language models (LLMs) have proven that a bottom-up reverse engineering of language at scale is a viable approach. However, due to their subsymbolic nature, LLMs do not provide us with an explainable model of how language works nor how we externalize the thoughts we contemplate in language. The idea of a bottom-up reverse engineering of language, which LLMs proved to be viable approach could however be done in a symbolic setting, as has been suggested previously going back to Frege. The obvious and ideal solution, therefore, would be to combine the advantages of a bottom-up reverse engineering approach with an explainable symbolic representation, as we have done in this paper. How the symbolic dimensions of word meanings we discussed in this paper are used in language understanding would be the subject of future work, but the reader can consult [17] for some suggestions on how this is accomplished.

**Figure 3.** Querying GPT-4 to discover plausible actions/relations and properties that can sensibly be said of a book, across three dimensions.

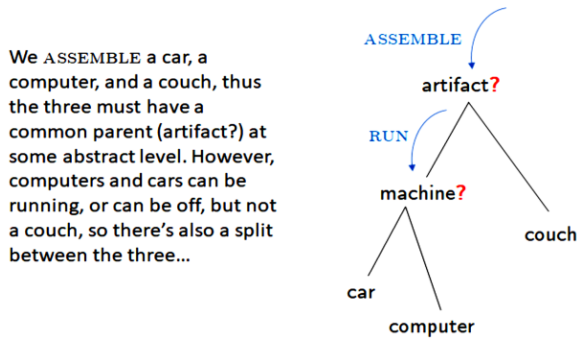

**Figure 4.** A computer, a car, and a couch can be assembled, so at some level of abstraction the three types must have a common parent. However, cars and computers, although not couches, RUN and can be ON or OFF and so they eventually must be in different branches.

Note that using this strategy we can also 'discover' the underlying ontology that seems to be implicit in our ordinary language. In